\pdfoutput=1
\newif\ifkouetsu
\kouetsufalse
\ifkouetsu
\documentclass[a4paper,12pt]{article}
\usepackage{txfonts}
\else
\documentclass[conference,a4paper]{./IEEEtran}
\fi
\newcommand{\ignore}[1]{}
\usepackage[pdftex]{graphicx}
\usepackage[cmex10]{amsmath}
\usepackage{bm}
\usepackage{amsxtra}
\usepackage{multirow}
\usepackage{cite}
\usepackage[small]{caption}
\usepackage[labelformat=simple,font+=footnotesize]{subcaption}

\ifkouetsu
\renewcommand{\baselinestretch}{2.4}
\else
\fi
\usepackage{newtxmath}
\begin{document}
 \bstctlcite{IEEEexample:BSTcontrol}
\title{%
\LARGE
Two-layer Lossless HDR Coding considering Histogram Sparseness with
Backward Compatibility to JPEG
}%
\ifkouetsu
\date{}
\else
\author{
\IEEEauthorblockN{Osamu WATANABE\dag}
\IEEEauthorblockA{\small Takushoku University\\
Dept. of Electronics \& Computer Systems\\
\dag Email: owatanab@es.takushoku-u.ac.jp}
\and
\IEEEauthorblockN{Hiroyuki KOBAYASHI\ddag}
\IEEEauthorblockA{\small%
Tokyo Metropolitan College of\\Industrial Technology\\
\ddag Email:hkob@metro-cit.ac.jp
}%
\and
\IEEEauthorblockN{Hitoshi KIYA\S}
\IEEEauthorblockA{\small Tokyo Metropolitan University\\
Faculty of Info. and Commun. Systems\\
\S Email: kiya@tmu.ac.jp}
}
\fi
\maketitle
\begin{abstract}
  An efficient two-layer coding method using the histogram packing
 technique with the backward compatibility to the legacy JPEG is proposed
 in this paper. The JPEG XT, which is the international standard to
 compress HDR images, adopts two-layer coding scheme for backward
 compatibility to the legacy JPEG. However, this two-layer coding structure does not
 give better lossless performance than the other existing single-layer
 coding methods for HDR
 images. Moreover, the JPEG XT has problems on
 determination of the lossless coding parameters;
 Finding appropriate
 combination of the parameter values is necessary to achieve good lossless
 performance.
 The histogram sparseness of HDR images is discussed and it is pointed
 out that the histogram packing technique considering the sparseness is
 able to improve the performance of lossless compression for HDR
 images and a novel two-layer coding with the histogram packing
 technique is proposed.
 The experimental results demonstrate that not only the proposed method
 has a better lossless compression
 performance than that of the JPEG XT, but also there is no need to determine
image-dependent parameter values for good compression
 performance in spite of having the backward
 compatibility to the well known legacy JPEG standard.
\end{abstract}
\section{Introduction}
The image compression method designed to provide coded data containing
high dynamic range content is highly expected to meet the rapid growth
of high dynamic range (HDR) image applications. Generally, HDR images
have much greater bit depth of pixel values and much wider color gamut\cite{ReinhardBook,8026195,ArtusiBook01,BADC11}.
These characteristic of HDR images are suitable for recording and/or
archiving the highly valuable contents, such as masterpieces of art.
For such a valuable content, HDR images should be losslessly encoded. In other
words, they should be compressed without any loss that is generated during
compression procedure.

Most of conventional image compression methods, however, could not
efficiently compress HDR image due to its greater bit depth and uncommon
pixel format including a floating point based pixel encoding.
Several methods have been proposed for compression of HDR
images\cite{Ward:2006:JBH:1185657.1185685,1528434,doi:10.1117/12.805315,6622714,4517823,6287996,6411962,6637869,7991151} and 
ISO/IEC JTC 1/SC 29/WG 1
(JPEG) has developed an international standard referred to as JPEG
XT\cite{ThomasPCS2013,Artusi2015,7426553,JPEGXT,7535096} for compression of an HDR image. JPEG XT has been designed to be backward compatible with
legacy JPEG\cite{JPEG-1} with two-layer coding; a base layer for
tone-mapped LDR image is compressed by the legacy JPEG encoder and an extension layer for residual data
consists of the result of subtraction between a decoded base layer image
and an original HDR image is compressed by the JPEG-like encoder. This
backward compatibility to legacy JPEG allows legacy applications and
existing toolchains to continue to operate on codestreams conforming to
JPEG XT.
Although this two-layer coding procedure makes it possible to
compress HDR images with the backward compatibility and the extension
layer contributes the improvement of the decoded image quality in lossy compression\cite{7532739}, its lossless compression
performance is not better than that of the other existing methods for HDR image
compression with single coding layer procedure.
The ISO/IEC IS 18477-8\cite{JPEGXTpart8},
which is known as the JPEG XT part 8,
makes it possible to encode HDR images losslessly
with such a two-layer coding procedure. In this part 8, it is
required to find a combination of the parameter values which gives a good
lossless compression performance. The combination could be dependent on
input HDR images. That is, finding the combination is required to
compress HDR images losslessly and efficiently.

In Refs.\cite{6288143,Minewaki2017,958146,1034993,988715,1040040,ELCVIA116,eusipco2017,6411962,6213328,6637869}, the sparseness of a histogram of an image is used
for efficient compression. `Sparse` histogram means that not all the bins in
a histogram are utilized. It is well known that a histogram of an HDR
image shows a tendency to be sparse\cite{6411962,6637869}.
 In Ref.\cite{6637869,KIYAJan2018,OsamuISCAS2018}, two-layer lossless coding
of HDR images has been proposed, however, methods in \cite{6637869,KIYAJan2018} are not backward
compatible with legacy JPEG. Ref.\cite{OsamuISCAS2018} has described the
JPEG compatible two-layer lossless coding with histogram packing,
however, the encoder for the extension layer has been fixed. In
addition, only the performance for HDR images with
floating point pixels has been examined and the effect of the histogram
packing has not been investigated.

This paper proposes a
new lossless two-layer method for both integer/floating-point HDR images with histogram packing and provides the
investigation of the effect of the histogram packing.
Codestreams
produced by the proposed method consist of two layers, i.e.
base layer and extension layer, where the base layer provides
low dynamic range (LDR) images mapped from HDR images by a
tone mapping operator (TMO), while the extension layer has the residual
information for reconstructing the original HDR images. For those
residual data, any lossless image encoders that can handle over 16 bits,
such as JPEG 2000 and JPEG XR, could be used. In
addition, the codestreams for the base layer are compatible with legacy JPEG
decoders. 
%
%

\section{Problems with JPEG XT lossless coding}
Because we
focus the lossless coding of HDR images with backward compatibility with
the legacy JPEG decoders,
the coding procedure of the JPEG XT part 8 is summarized and then
the problem with it is described.
\begin{figure}[tb]
 \centering
 \includegraphics[width=1.0\columnwidth]{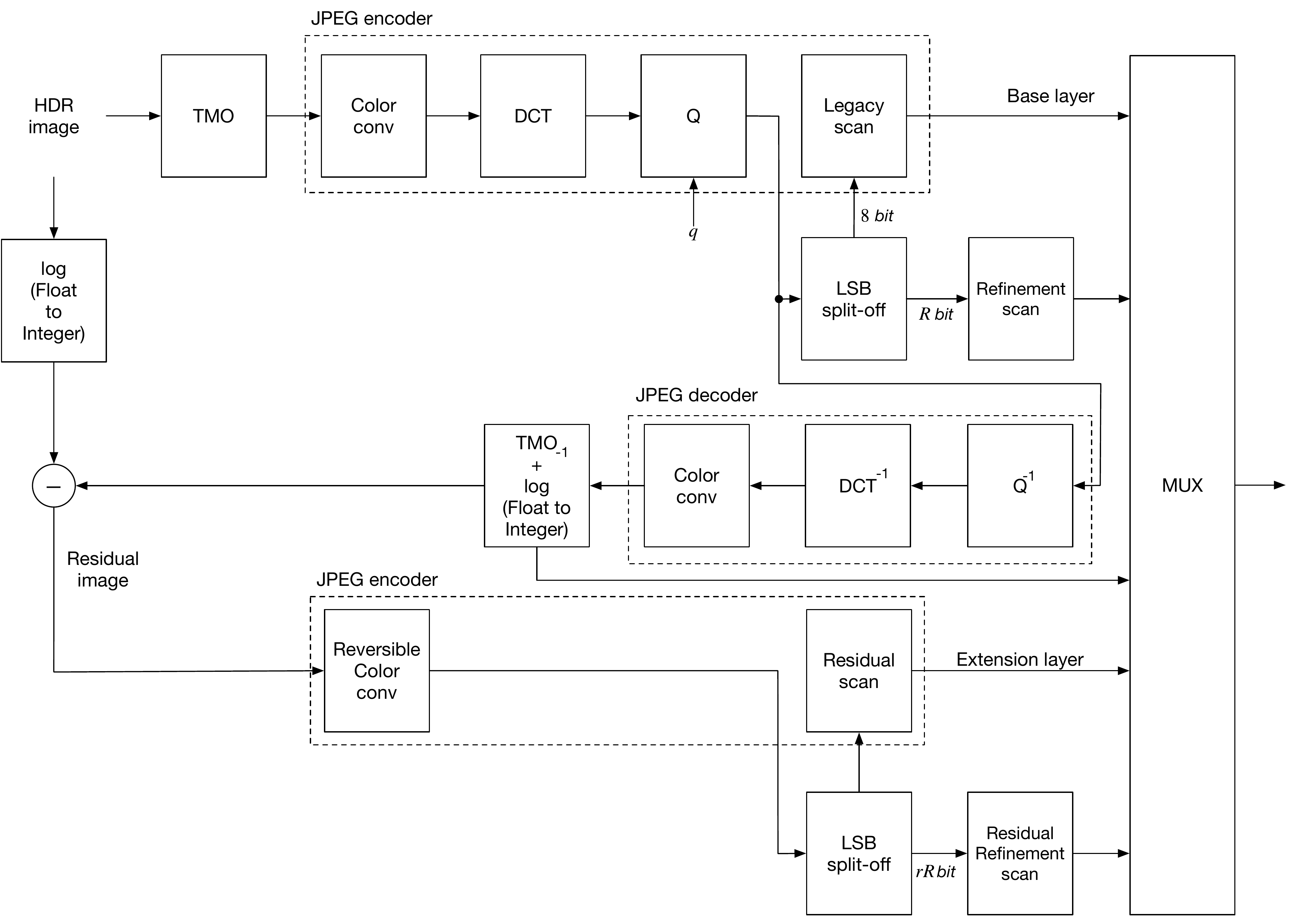}
 \caption{Blockdiagram of JPEG XT part 8 encoder: 'TMO' means tone
 mapping operator. $Q$ and $Q^{-1}$ are quantization and inverse
 quantization, respectively. $q$ is parameter to control quality of
 decoded base layer (LDR) image.}
 \label{XT_C_enc}
\end{figure}
%
The blockdiagram of
the part 8 encoder is shown in Fig.\ref{XT_C_enc}.
Although the pixel values of HDR images are often represented with
floating point numbers, these floating point numbers are re-interpreted as
integer number with IEEE floating point representation\cite{4610935,595279}. This
representation is exactly invertible\cite{ThomasXT} and makes it
possible to compress HDR images losslessly.

For lossless compression of HDR images, it is required to determine the
values of several parameters. The first parameter is $q$, which
controls decoded image quality of base layer. The higher $q$ gives the
better quality. The second parameter $R$ is the number of bits used for
refinement scan. The refinement scan is used to improve precision of DCT
coefficients up to 12 bit. Thus the valid range of $R$ is from $0$ to $4$.
The third parameter is $rR$. The $rR$ is the number of bits used for
residual refinement scan. In lossless coding procedure, the $rR$
is considered as the control factor for the amount of coded data
included in the residual data of the extension layer.

\begin{figure}[b]
 \centering
 \includegraphics[width=0.8\columnwidth]{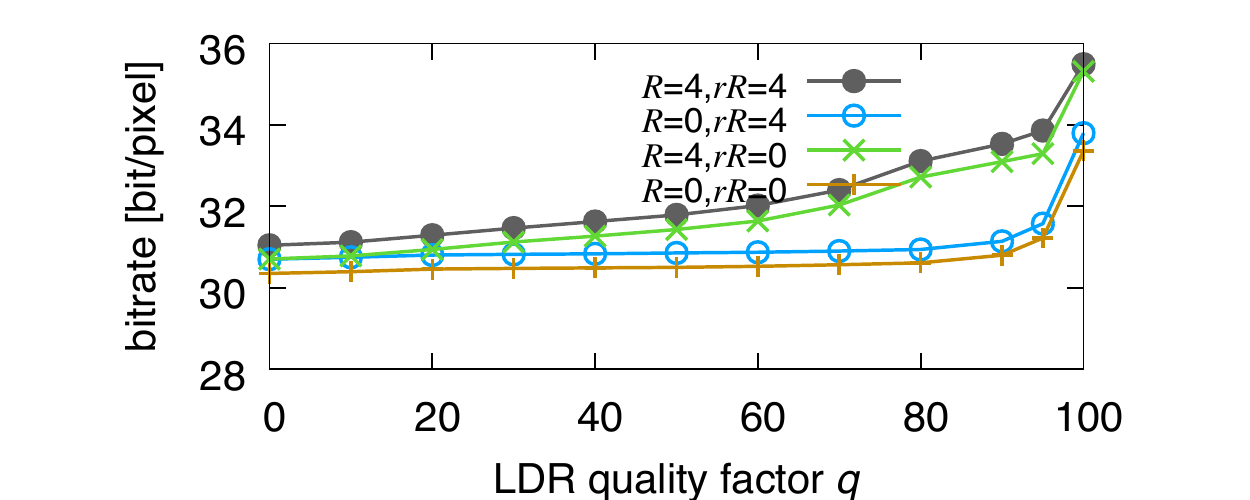}
 \caption{Bitrate of lossless compressed HDR image (MtTamWest) by JPEG XT}
 \label{XTwithRef}
\end{figure}

To achieve good lossless compression performance, the values of the
parameters, $q$, $R$ and $rR$ should be carefully determined. Figure
\ref{XTwithRef} shows the result of lossless compression of an HDR image
by the part 8 with $q=0$ to $100$, $R=4$ and $rR=0$.
Clearly, we can see there is a certain variation in the coding
performance. Note that it has been confirmed that the optimal values of the
parameters which give the best performance is image-dependent.

\section{Proposed method}
A method using the histogram packing technique with
the two-layer coding having the backward compatibility to the legacy
JPEG for base layer is described in this section.
\subsection{Histogram sparseness of residual data}
\label{IIB}
\begin{figure}[t]
 \centering
\subcaptionbox{memorial}{
 \includegraphics[width=0.465\columnwidth]{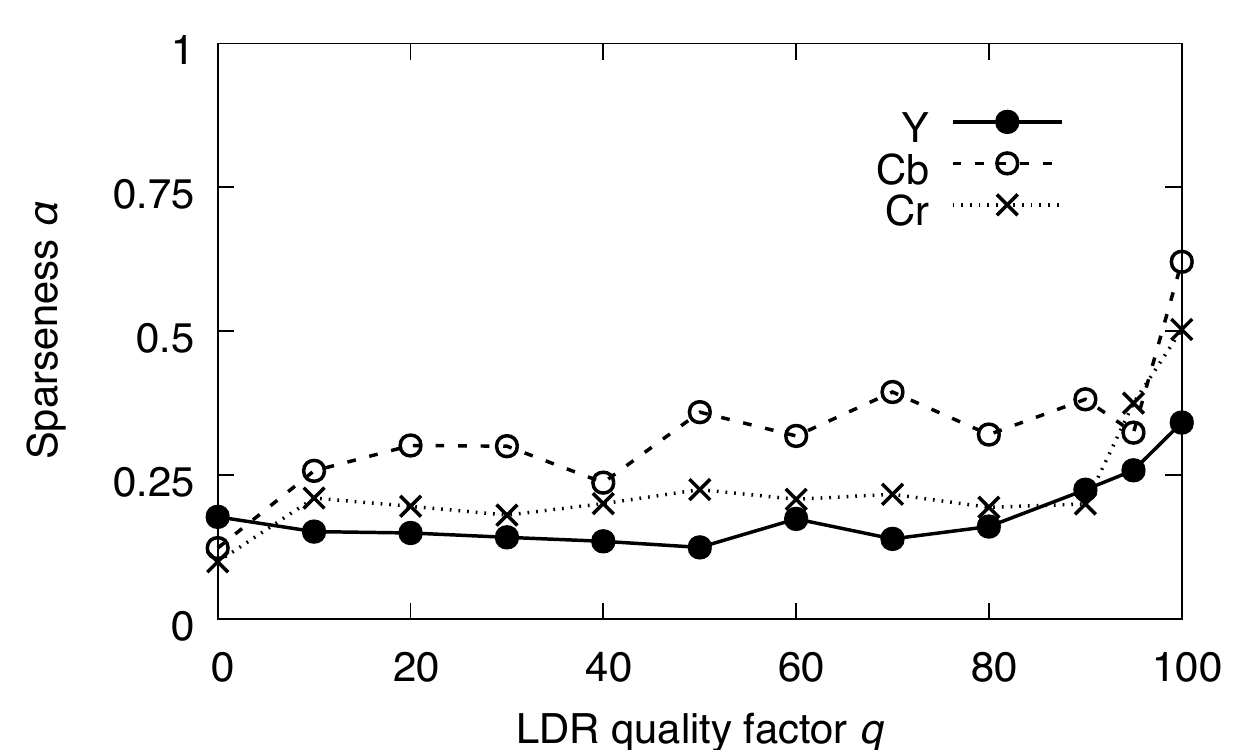}
 }
 \subcaptionbox{MtTamWest}{
 \includegraphics[width=0.465\columnwidth]{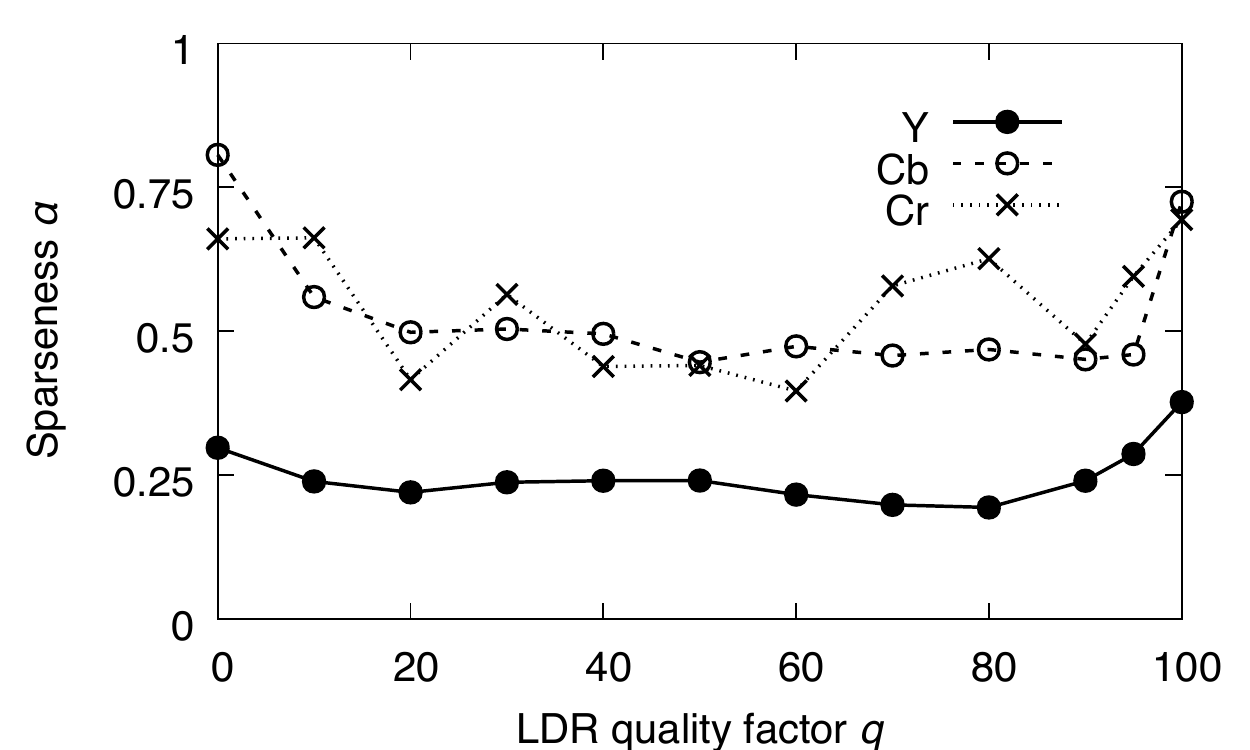}
 }
 \caption{Histogram sparseness of residual data (16bit floating-point)}
 \label{histsparse_float}
\end{figure}
\begin{figure}[t]
 \centering
\subcaptionbox{Cafe}{
 \includegraphics[width=0.465\columnwidth]{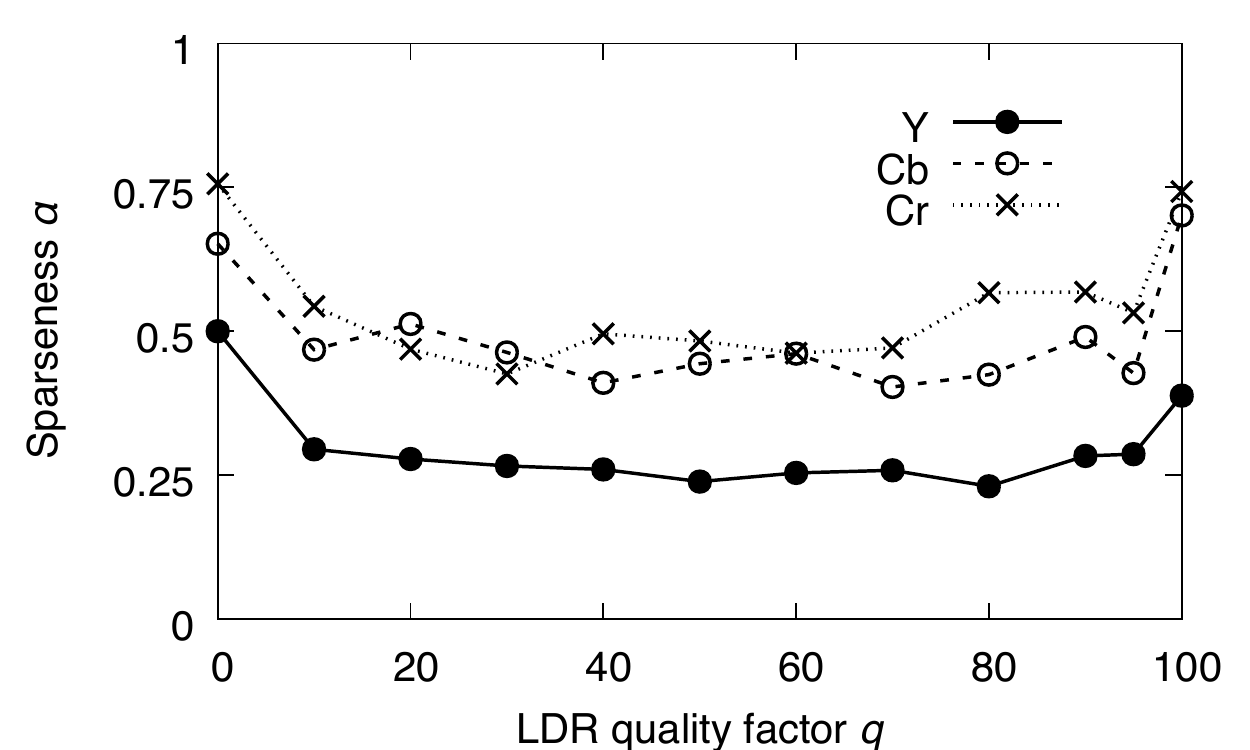}
 }
 \subcaptionbox{Books}{
 \includegraphics[width=0.465\columnwidth]{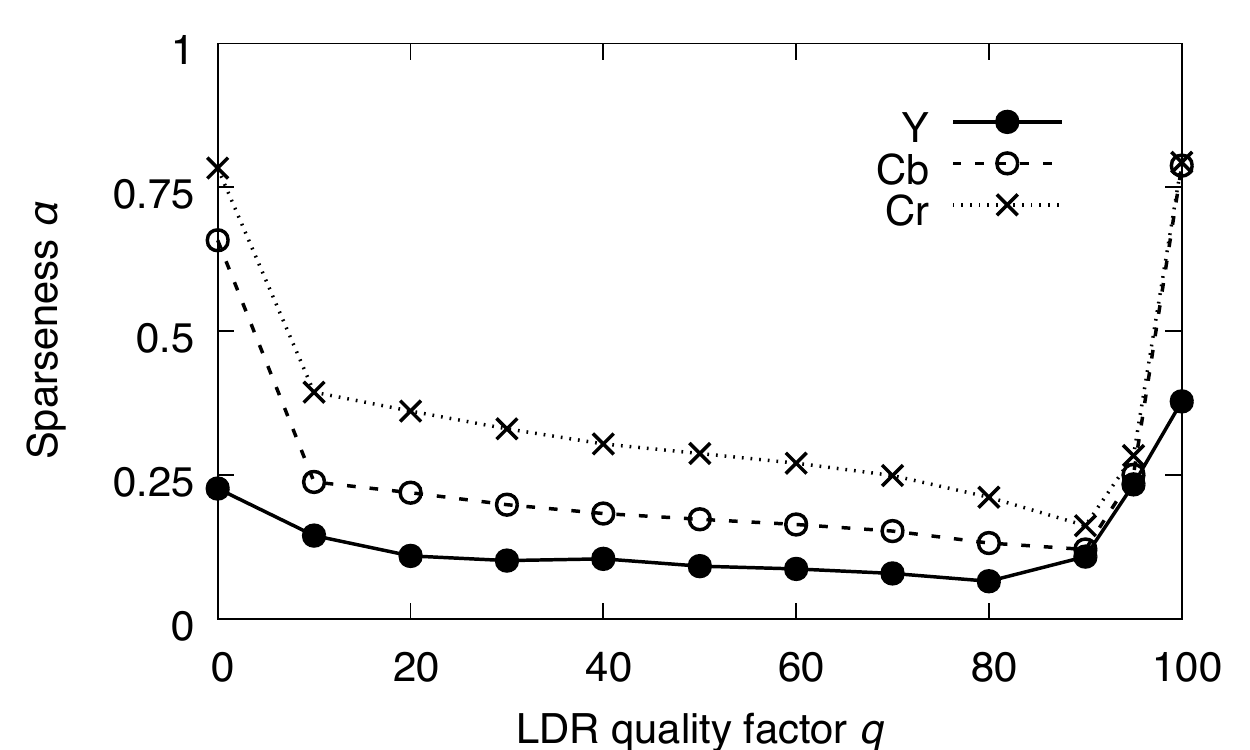}
 }
 \caption{Histogram sparseness of residual data (16bit integer)}
 \label{histsparse_int}
\end{figure}
HDR images often have sparse histograms due to its high dynamic range of
pixel values\cite{6637869}. Moreover, the histograms of the residual data in the
two-layer coding in the part 8 are also sparse after subtraction of
LDR data in the base layer. In this paper,
this histogram sparseness is denoted as $\alpha$ and 
defined by
\begin{align}
 \alpha &= \frac{|X|}{\max(x)_{x\in X} - \min(x)_{x\in X} + 1}\\
 X &= \{x|H(x)\neq 0\}
\end{align}
where $H(x)$ denotes the histogram of a pixel value $x$, and $|X|$
denotes the total number of all the elements of a set $X$. The range of
$\alpha$ is $0\leq \alpha \leq 1$ and the greater $\alpha$ means the
 sparser histogram.
Figure \ref{histsparse_float} and \ref{histsparse_int} show the {\it `sparseness'} of the residual data of
two types of HDR images having floating-point and integer pixel values. The remarks from these figures are
summarized as follows.
\begin{itemize}
\item The sparseness depends on images and the quality factor $q$ for base layer. 
\item The histogram of residual data tends to be sparse, especially, the value
of $\alpha$ in chroma component is higher than that in luminance.
\end{itemize}
For image signals having such a sparseness, it is well known that the
histogram packing technique improves lossless
compression performance\cite{958146,1034993,988715,1040040,ELCVIA116,eusipco2017}. The main idea of the proposed
method is to combine the two-layer coding structure with the histogram
packing technique.

\subsection{Histogram packing}
  \begin{figure}[tb]
   \includegraphics[width=1.0\linewidth]{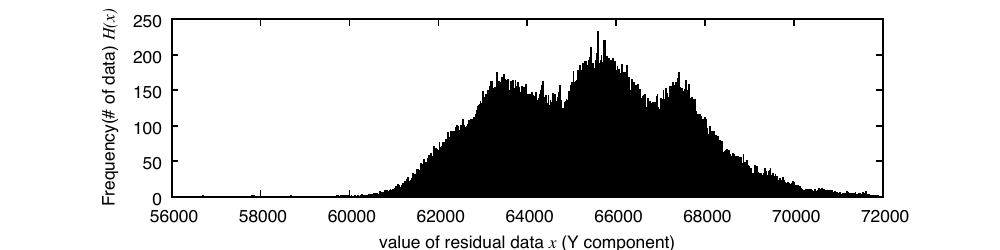}
   \caption{Histogram of residual data (Y component of 'BloomingGorse2',
   LDR $q=50$, Sparseness $\alpha=0.374$)}
   \label{partialHisto}
  \end{figure}
  \begin{figure}[tb]
   \subcaptionbox{Index image (after histogram packing)\label{idximg}}[0.49\linewidth]{
   \includegraphics[width=0.47\linewidth]{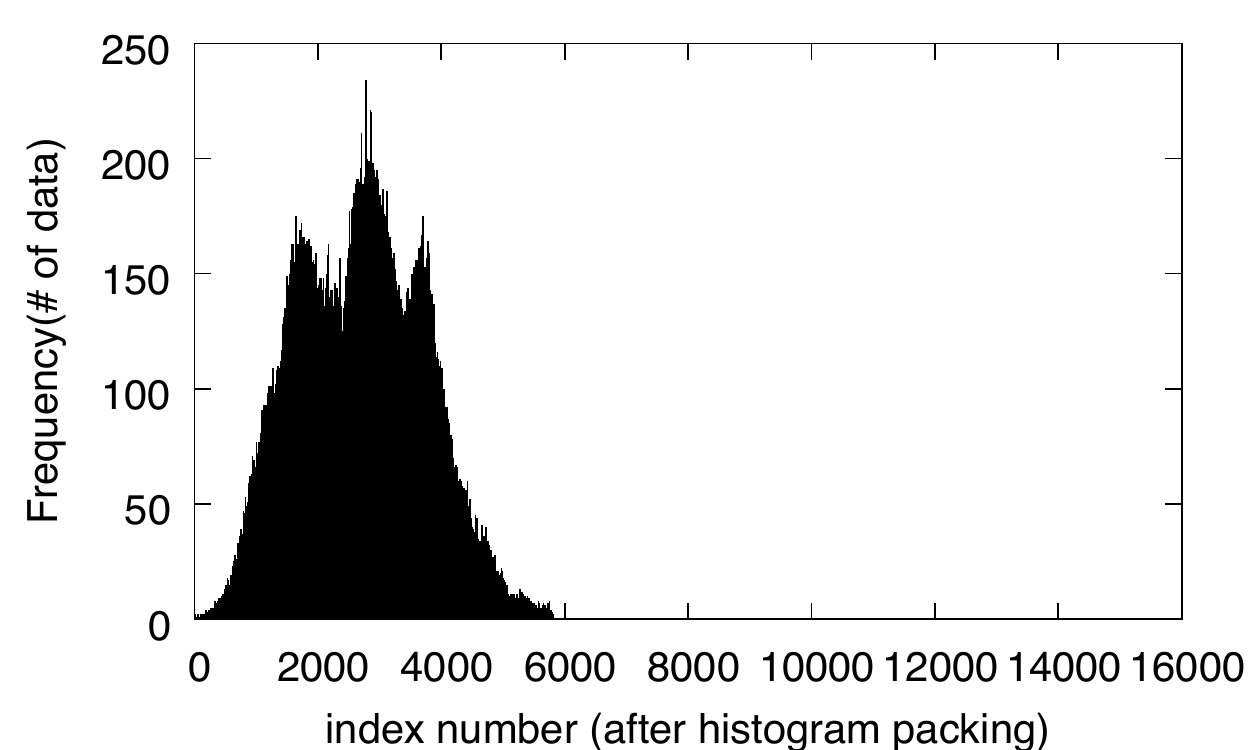}
   }
   \subcaptionbox{Unpacking table\label{utable}}[0.49\linewidth]{
   \includegraphics[width=0.47\linewidth]{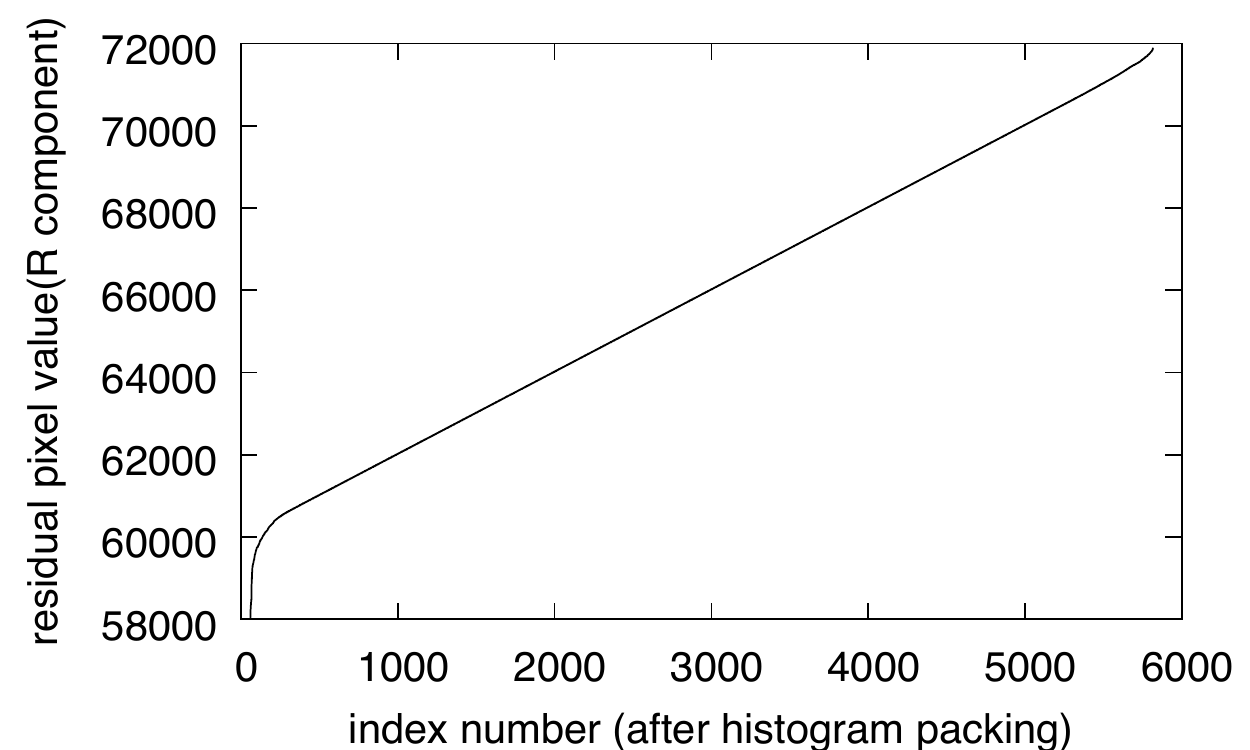}
   }
  \caption{Histogram of index image and unpacking table (Y
   component of 'BloomingGorse2', LDR $q=50$)}
   \label{packedIdxTbl}
  \end{figure}
In previous subsection, it has been noted that the histograms of the residual data tend to be
sparse and the reduction of the sparseness is effective to improve
lossless coding performance. Figure \ref{partialHisto} shows a histogram
$H(x)$ for Y component in the residual data of an HDR image
'BloomingGorse2.' Horizontal axis denote the pixel values $x$ in integer
number with IEEE floating point representation.
After histogram
packing, a histogram-packed image is obtained. In this paper, this
histogram-packed image is referred to as 'index image.' $H(x)$ for the
index image is shown in Fig. \ref{idximg} and it is clearly considered
to be dense. The index image is compressed by the lossless image encoder.
The unpacking table, which is necessary to perform inverse histogram
packing, is illustrated in Fig. \ref{utable}. Obviously, it is
considered as one-to-one correspondence function and monotonically increasing.
That is, this table is DPCM effective．
  \subsection{Encoder structure}
\begin{figure}[tb]
 \centering
 \includegraphics[width=1.0\columnwidth]{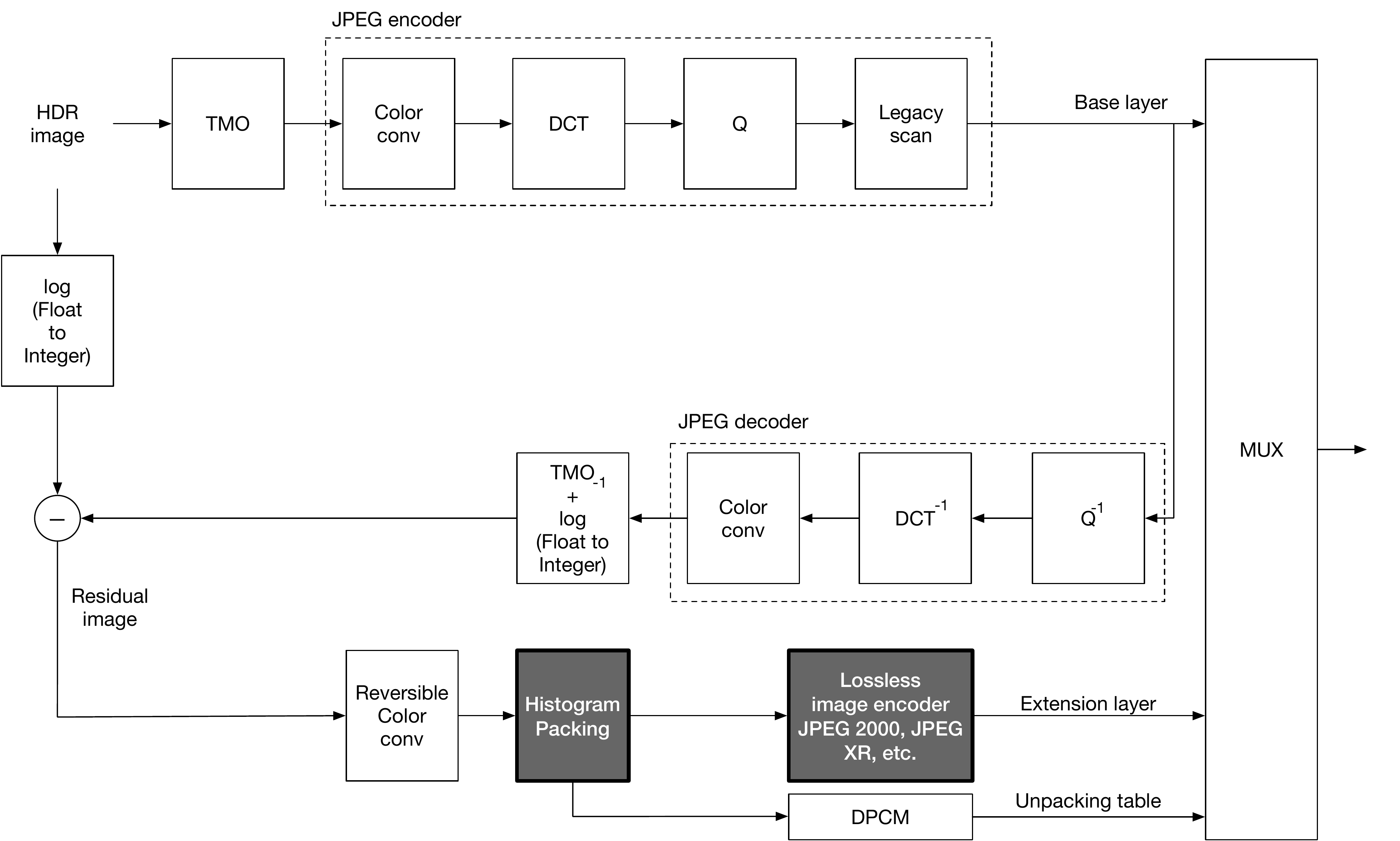}
 \caption{Blockdiagram of proposed two-layer lossless encoder}
 \label{proposed_enc}
\end{figure}
The structure of the proposed lossless two-layer coding is illustrated in
Fig.\ref{proposed_enc}. The coding-path to generate a base layer, which
is backward compatible with the legacy JPEG, is the exactly same as the
JPEG XT part 8. Note the refinement scan for the base layer is not
used. Therefore, the value of $R$ is set to zero.

For the extension layer, which consists of the residual data generated
by subtracting decoded base layer from the original HDR image, the
coding procedure after color space conversion from RGB to YCbCr is
different from the part 8 encoder.
The histogram of each color
component of the color converted residual data is analyzed and packed by
using the histogram packing technique. Then, the packed residual data is
compressed by the lossless image encoder, such as the JPEG 2000, JPEG XR
 etc. After the
subtraction described above, the residual data for each color component could
have 17 bit integers. This over 16 bit in the bit-depth is the reason
for using such lossless encoders as the JPEG 2000 and the JPEG XR
because they are able to accept up to 32 bit integer pixel value
per component\cite{1528434}.
For the inverse operation of the histogram packing, unpacking table is
sent to the decoder. The unpacking table is one-to-one correspondence
function between the packed index value and the original pixel
value. Since this is monotonically increasing, DPCM and bzip2
compression are performed to
reduce the data amount of the table.

The base layer which is compatible with the legacy JPEG, the extension
layer consists of the lossless JPEG 2000 or JPEG XR codestream, and the compressed
unpacking table are multiplexed into single codestream and it is sent to the
decoder.

\section{Experimental results}
To verify the effectiveness of the proposed method, the lossless compression performance in terms of
bitrate of the generated codesrtreams was evaluated and compared with
that of the JPEG XT part 8.
\subsection{Conditions}
Images having both floating-point and integer pixel values were selected for
the experiments. For floating-point images, four of images common to HDR related
experiments were collected.
For integer images, four of the ITE test
images\cite{ITEIMG} were used. The specifications of these test images are summarized in
Table \ref{testimages}. 
Although some of floating-point images have full precision float value for their
pixel value, we have converted the values into half precision float because
the JPEG XT encoder only accepts half precision floating point pixels as
its inputs.
Note that image names are all represented by the index shown in
Table\ref{testimages}. The first character of the index means the type
of pixel values; ``f'' is for floating-point and ``i'' is for
integer. The second number stands for each image's name.
For the JPEG XT part 8 encoder, the reference software\cite{JPEGXTpart5,JPEGXTSOFT} available from
the JPEG committee was used. For the proposed method, the modified
encoder of the reference software, whose coding path for the residual
data was changed to have the histogram packing and JPEG 2000/JPEG XR encoder,
was used. The Kakadu software\cite{Kakadu} and the
reference software of the JPEG XR\cite{JPEGXRSOFT} were used as those encoders that were used to compress the
histogram-packed residual data.
The lossless performances of the proposed method and the JPEG XT part 8 were evaluated with several values of $q$ (quality factor of LDR
image) and $R$ (number of refinement bits for base layer). For the JPEG
XT, another parameter, the effect of $rR$ (number of refinement bits for extension
layer), was also evaluated.
 \begin{table}[tb]
  \centering
  \caption{Test images (bpp means bit-depth per component): All images have three color components in RGB color space}
\label{testimages}
   \begin{tabular}{c|c|c|c|c}
    \hline
   Pixel value type & Index & Name & bpp & Size \\\hline
  \multirow{4}{*}{Floating-point}& f1 & memorial & 16 & 512$\times$768\\
  & f2& Blooming Gorse2& 16 & 4288$\times$2848\\
  & f3& MtTamWest & 16 & 1214$\times$732\\
  & f4& Desk& 16 &644$\times$874\\ \hline
  \multirow{4}{*}{Integer}& i1 & Book & 12 & 3840$\times$2160\\
  & i2& Kimono & 12 & 3840$\times$2160\\
  & i3& Moss & 12 &3840$\times$2160\\
  & i4& MusicBox& 12 & 3840$\times$2160\\ \hline
   \end{tabular}
 \end{table}
\subsection{Results and remarks}
\subsubsection{Lossless performance}
 \begin{figure}[t]
  \centering
  \subcaptionbox{f1\label{res_f1}}{
  \includegraphics[width=0.465\linewidth]{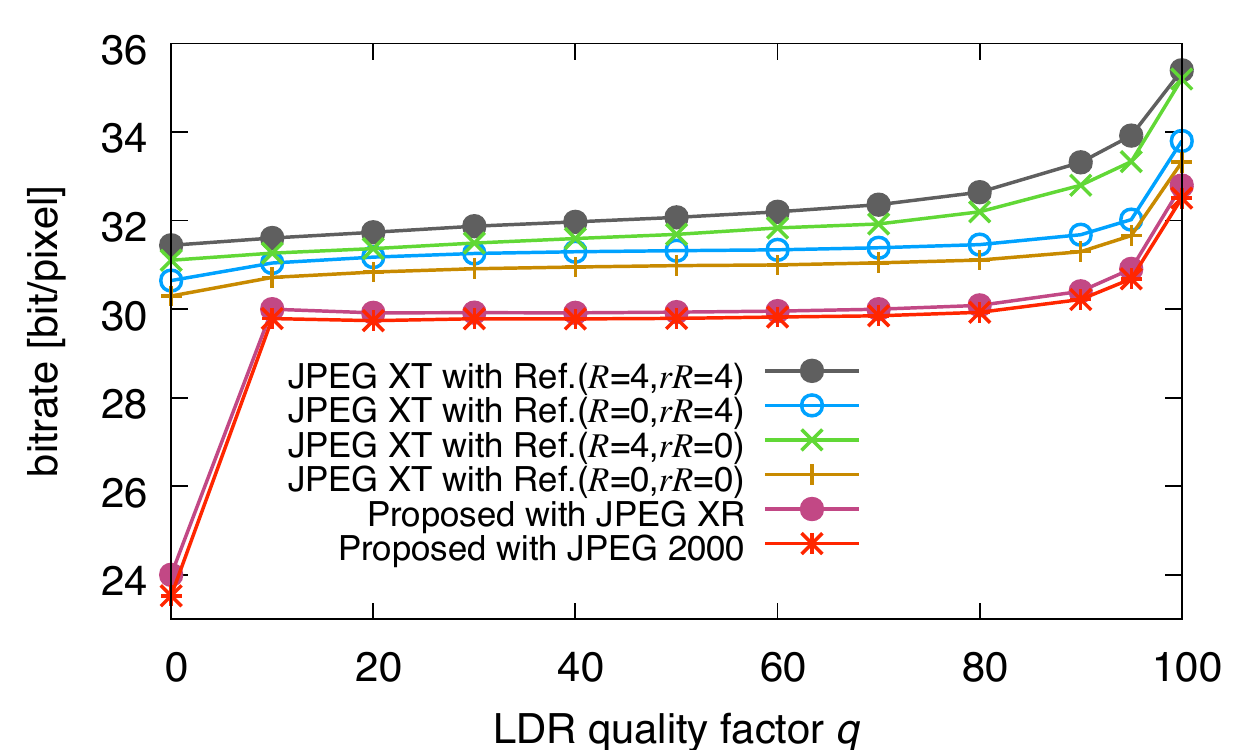}
  }
  \subcaptionbox{f2\label{res_f2}}{
  \includegraphics[width=0.465\linewidth]{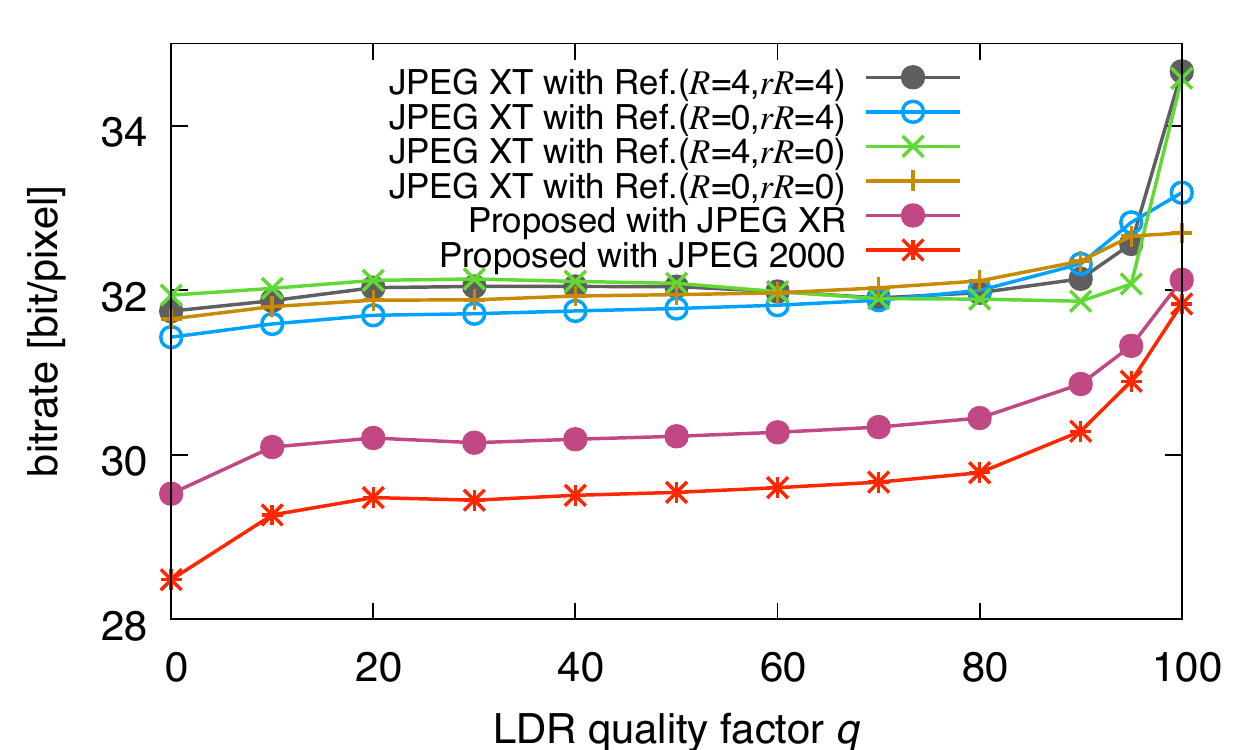}
  }
  \subcaptionbox{f3\label{res_f3}}{
  \includegraphics[width=0.465\linewidth]{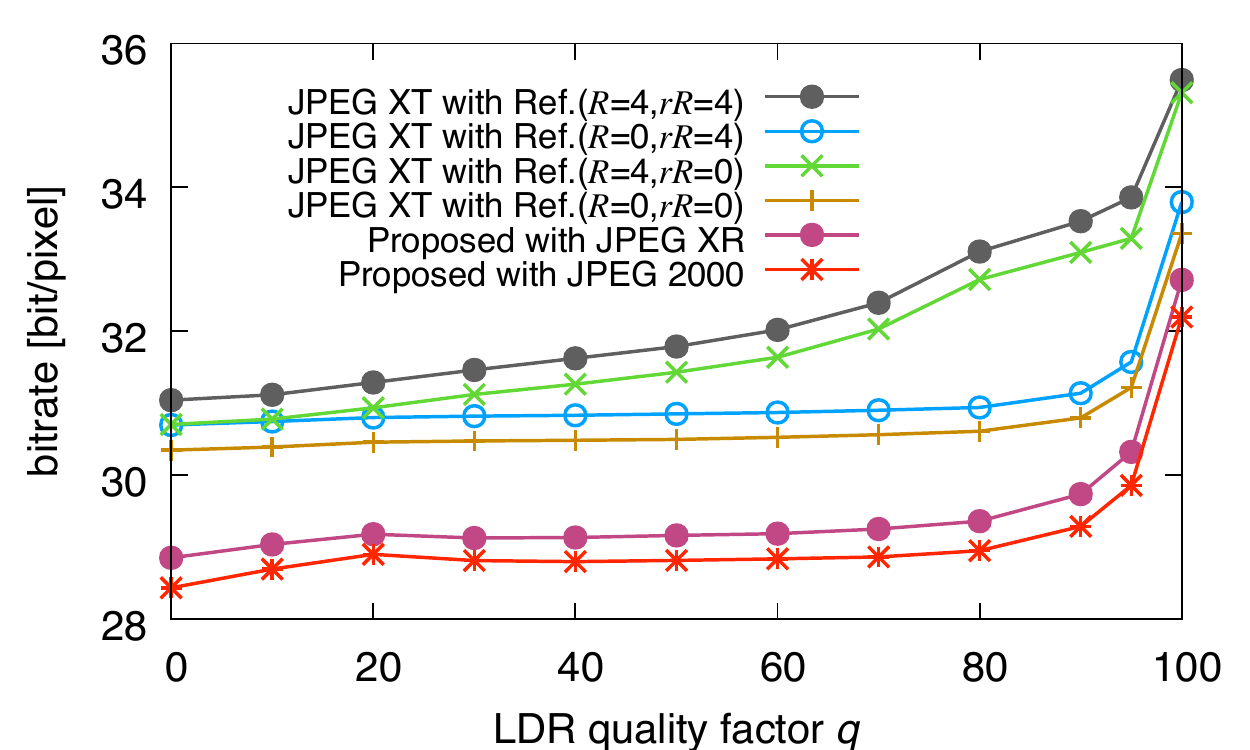}
  }
  \subcaptionbox{f4\label{res_f4}}{
  \includegraphics[width=0.465\linewidth]{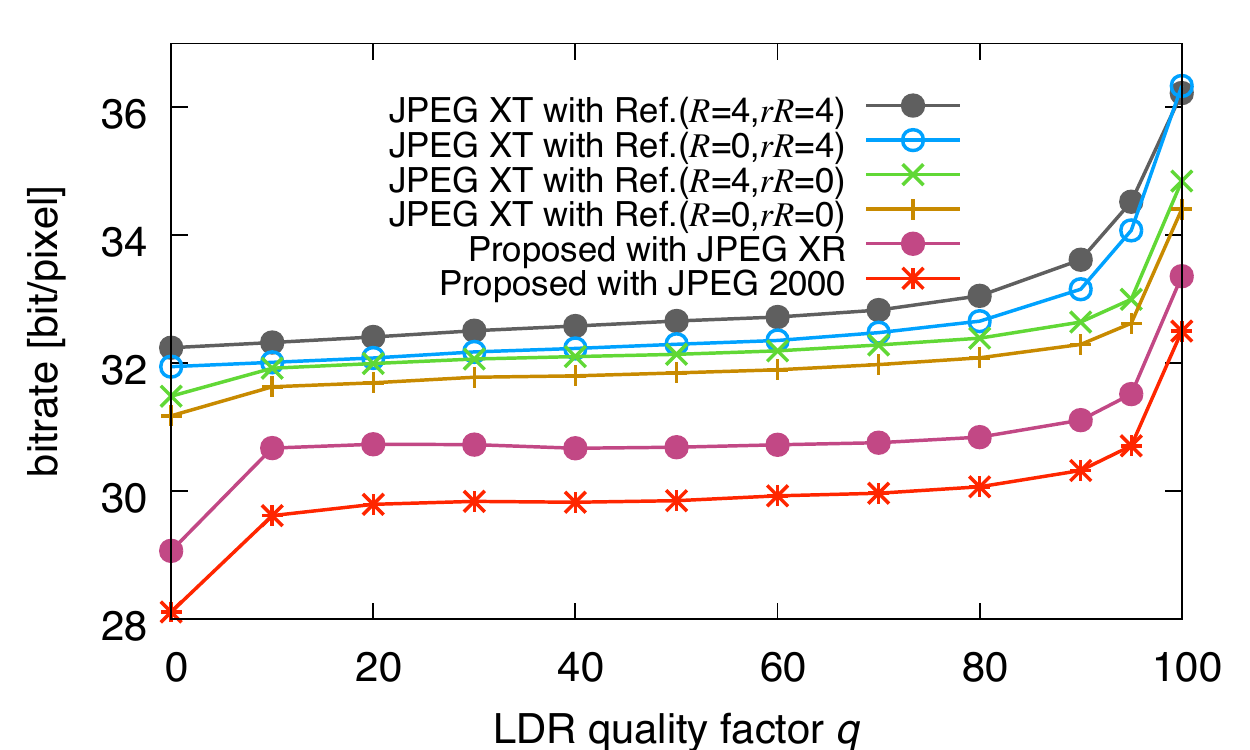}
  }
  \caption{Bitrates of lossless compressed image (float): image names are represented by index (see Table. \ref{testimages}.)}
  \label{floating_results}
 \end{figure}
  \begin{figure}[t]
  \centering
  \subcaptionbox{i1\label{res_i1}}{
  \includegraphics[width=0.465\linewidth]{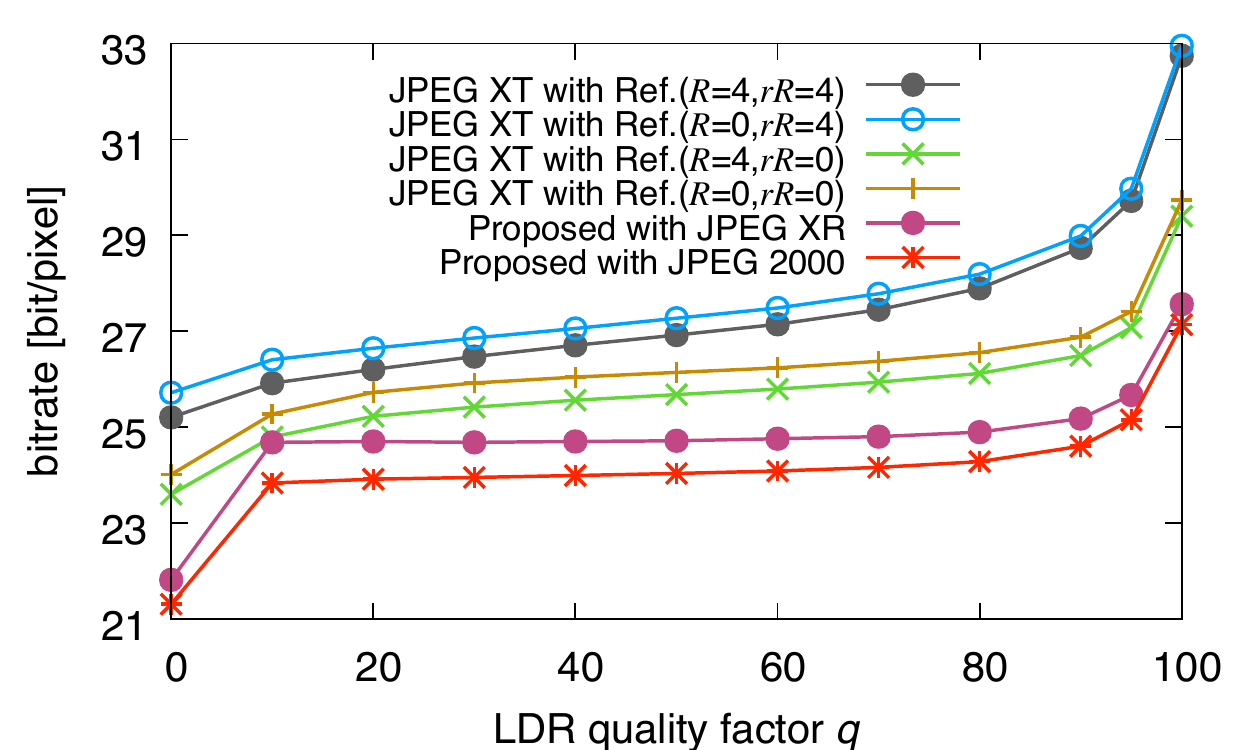}
  }
  \subcaptionbox{i2\label{res_i2}}{
  \includegraphics[width=0.465\linewidth]{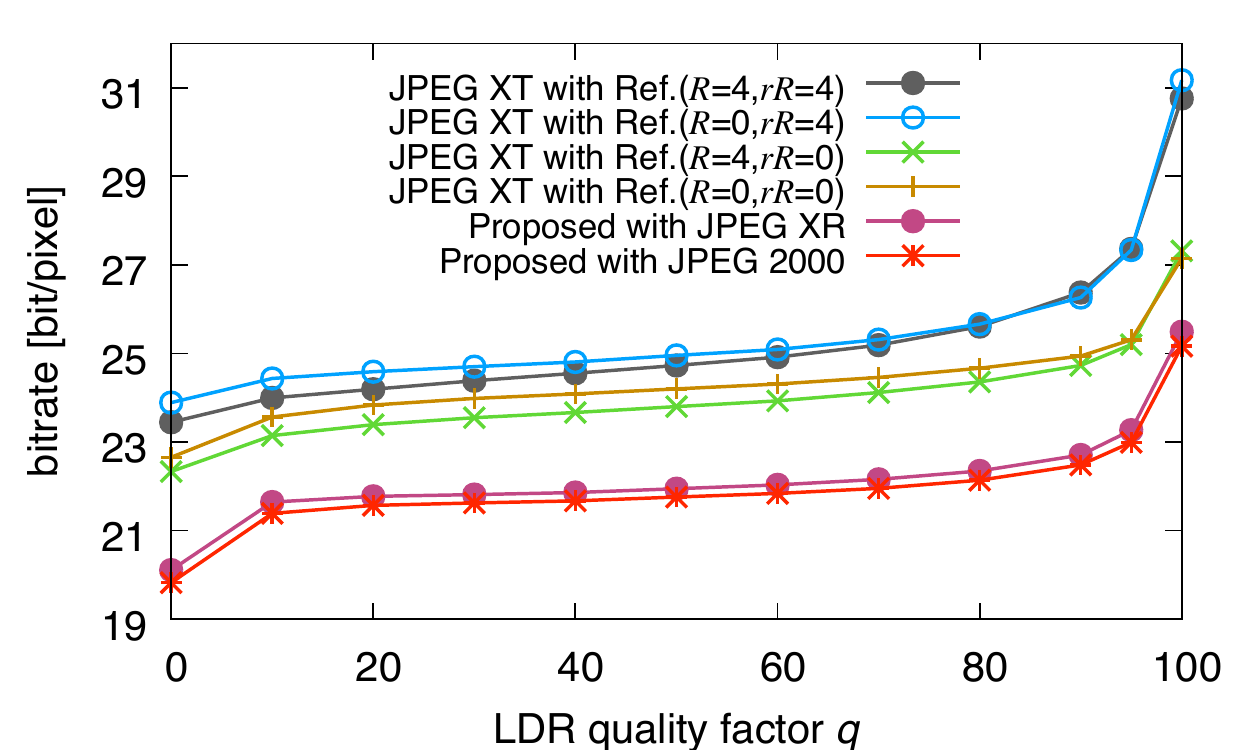}
  }
  \subcaptionbox{i3\label{res_i3}}{
  \includegraphics[width=0.465\linewidth]{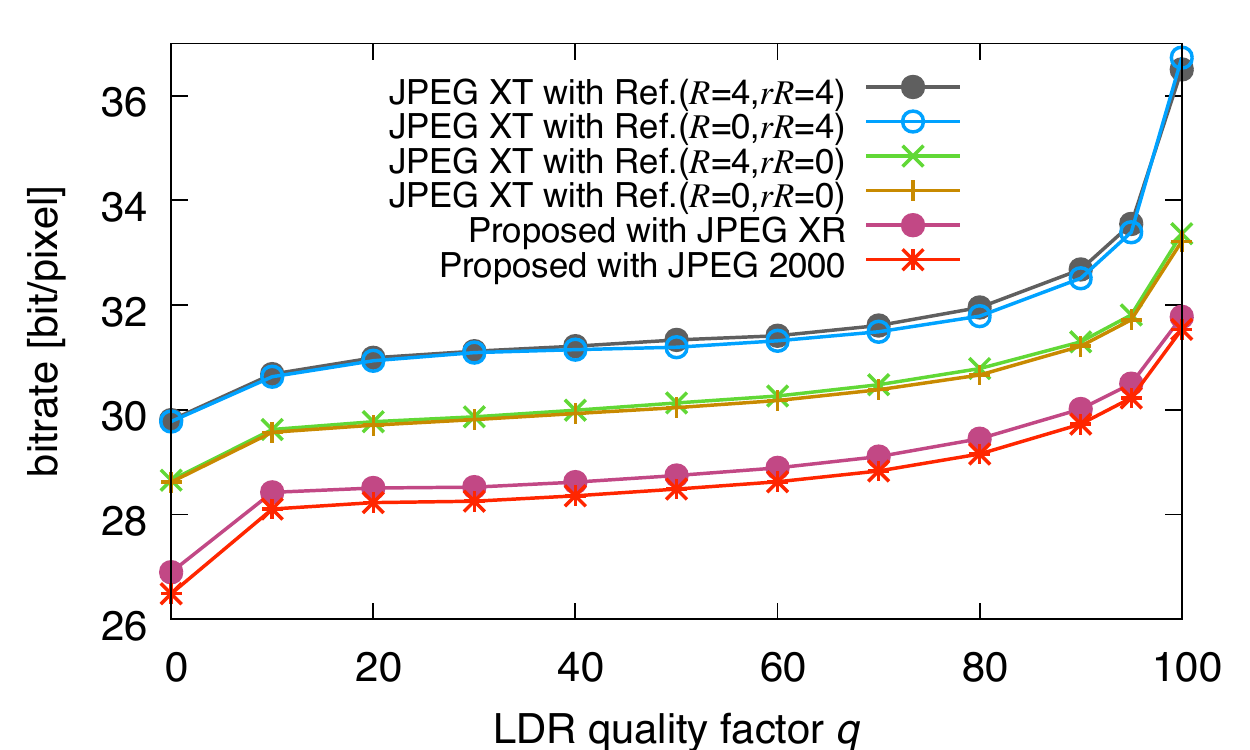}
  }
  \subcaptionbox{i4\label{res_i4}}{
  \includegraphics[width=0.465\linewidth]{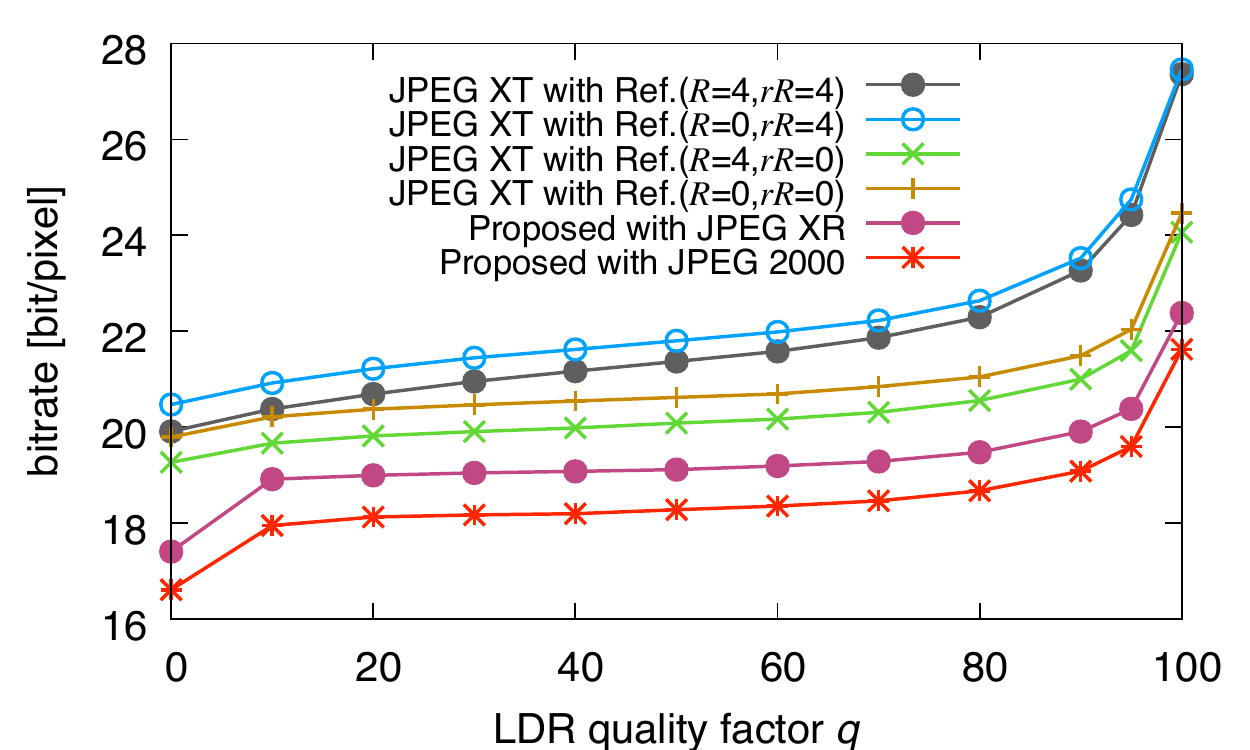}
  }
  \caption{Bitrates of lossless compressed image (integer): image names are represented by index (see Table. \ref{testimages}.)}
  \label{integer_results}
  \end{figure}
Figures \ref{floating_results} and \ref{integer_results} show the results of lossless bitrate with the
proposed method with the different LDR quality $q$ and those with the
JPEG XT part 8 with the different parameter values of
$q$, $R$ and $rR$.
From these results, it is clearly confirmed that the
proposed method shows the best lossless performance regardless of
images and those pixel value types, the values of $q$.
%
Figures \ref{floating_results} and \ref{integer_results} show the results of lossless bitrate with the
proposed method with the different LDR quality $q$ and those with the
JPEG XT part 8 with the different parameter values of
$q$, $R$ and $rR$.
From these results, it is clearly confirmed that the
results of the proposed method show the better lossless performance regardless of
images and those pixel value types, the values of LDR quality $q$. 
It is worth noting that the values of $R$ and/or $rR$ should be
carefully determined for the JPEG XT. For example, in Fig.\ \ref{res_f2}, the
bitrates of JPEG XT with the combination of the refinement parameters $(R=4, rR=0)$ from $q=0$
to $q=60$ are better than those with the other combinations, however,
its bitrates over $q=80$ are the worst. This means
the best combination of $R$ and $rR$ depends on the LDR quality $q$ and
the input image.
On the other hand, the proposed method with the JPEG 2000 encoder gives the first best
performance. The second best is the result of proposed method with the
JPEG XR encoder. Although there is some difference between the results,
those two types of the proposed method give the lower bitrate than those
obtained by the JPEG XT encoder, even though there is no dependency on the LDR
$q$ and the input image. Note that it is verified that the ratios of the
data amount for the unpacking table to the total bitrate are less than 0.4\% at maximum.
\subsubsection{Effect of Histogram packing}
\begin{figure}[tb]
 \centering
 \subcaptionbox{f1}{
 \includegraphics[width=0.465\columnwidth]{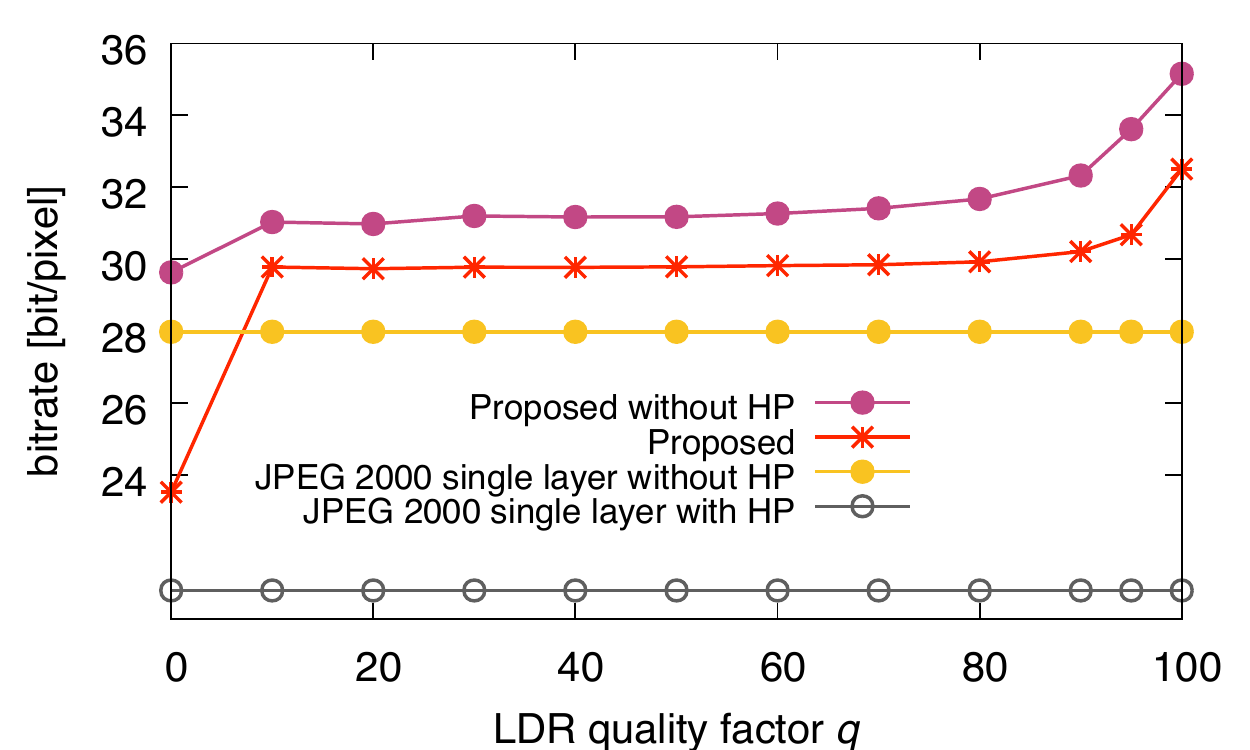}
 }
 \subcaptionbox{f2}{
 \includegraphics[width=0.465\columnwidth]{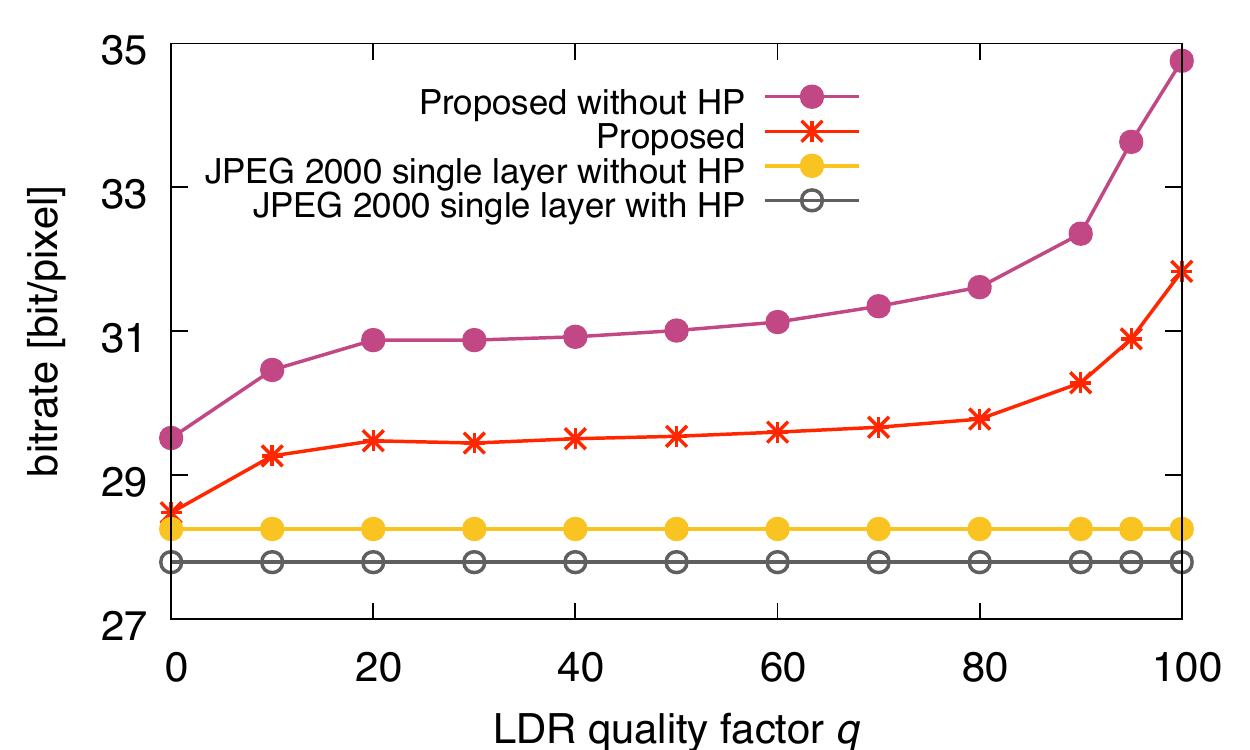}
 }
 \subcaptionbox{f3}{
 \includegraphics[width=0.465\columnwidth]{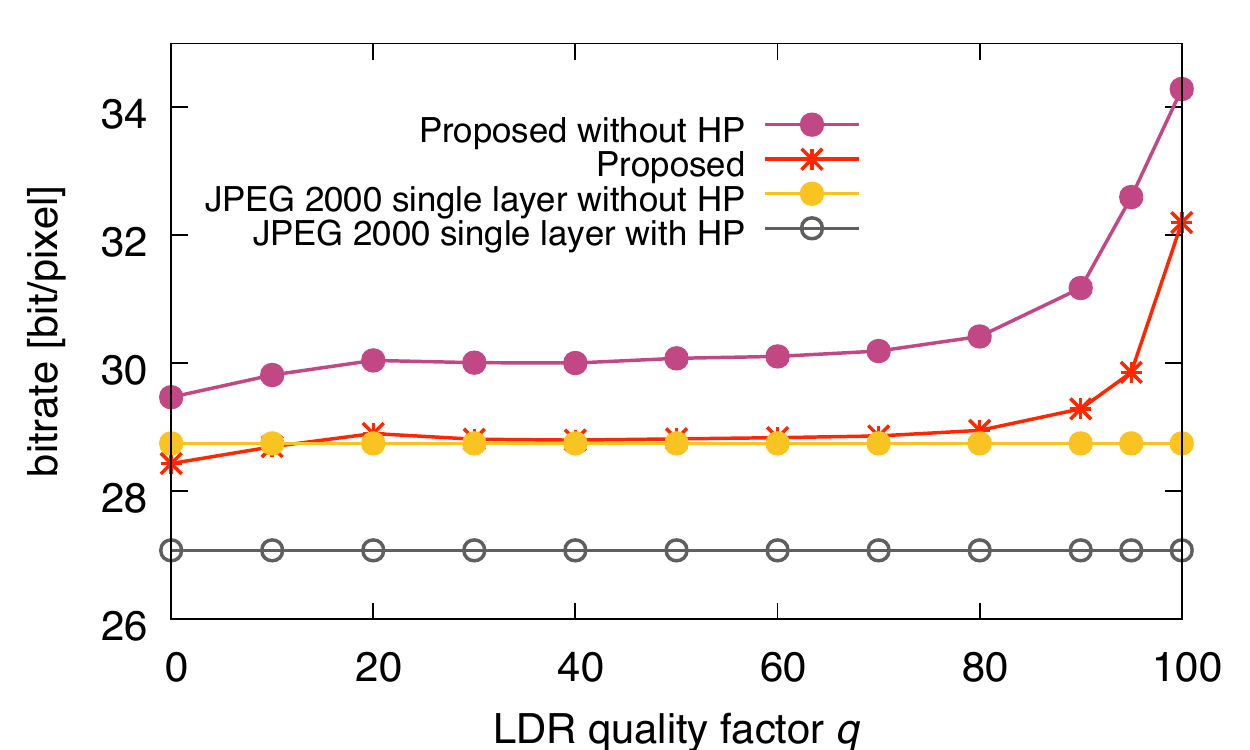}
 }
 \subcaptionbox{f4}{
 \includegraphics[width=0.465\columnwidth]{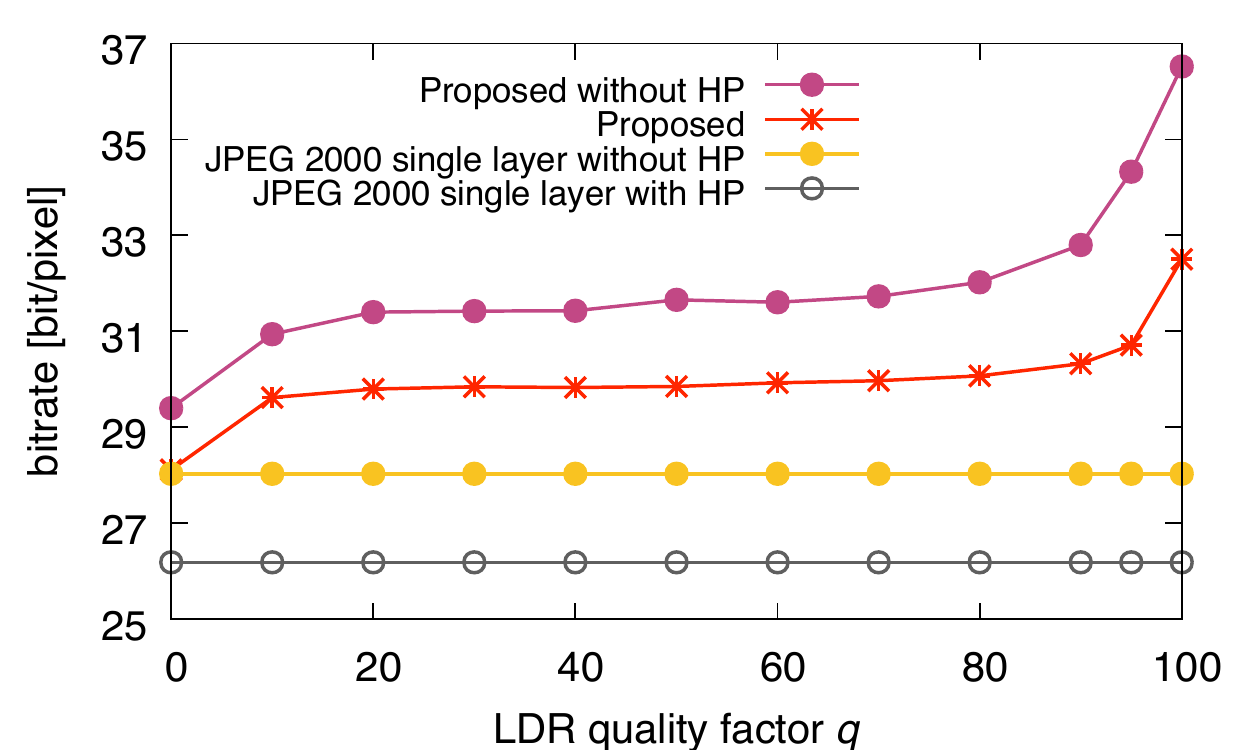}
 }
 \caption{
 Effect of histogram packing: red lines are lossless bitrates with
 hitogram packing and purple lines are that without histogram packing.}
 \label{comp_hist_nonhist_1_2}
\end{figure}
To verify the effectiveness of histogram packing for lossless
compression, lossless bitrates in single layer JPEG 2000 with/without
histogram packing and two layer coding without histogram packing and
the proposed method were examined.
Figure \ref{comp_hist_nonhist_1_2} shows those lossless bitrates for
floating-point HDR images. The
remarks from this figure are two: a) 
Clearly, lossless bitrates with histogram packing are smaller than those
without histogram packing. It has been confirmed that the use of JPEG XR encoder instead of
JPEG 2000 provide the same trend and the results for integer HDR images
denote the same tendency. b) The histogram sparseness is 
 effective
to improve lossless compression performance of HDR images having sparse
histogram, regardless of whether the encoder structure is two or single layer.

 \section{Conclusions}
 A novel method using the histogram packing technique with the two-layer coding having the
 backward compatibility with the legacy JPEG for base layer has been proposed
 in this paper.
 The histogram packing technique has been used to improve the performance of
 lossless compression for HDR images that have the histogram sparseness.
 The experimental results in terms of lossless bitrate have demonstrated that the proposed
 method has a higher compression performance than that of the JPEG XT
 Part 8. Unlike the JPEG XT Part 8, there is no need to determine
 image-dependent values of the coding parameters to achieve good compression
 performance. Moreover, the base layer produced by the proposed method has the backward
 compatibility with the legacy JPEG standard, which is one of the most
 spread image format.
 
\bibliographystyle{IEEEtran}
\bibliography{./IEEEabrv,./refs}
\label{sec:ref}
\end{document}